\documentclass[10pt,letterpaper]{article}

\usepackage[margin=1.2in]{geometry}
\usepackage{times}
\usepackage{epsfig}
\usepackage{graphicx}
\usepackage{subcaption}
\usepackage{amsmath}
\usepackage{amssymb}
\usepackage{mathtools}

\usepackage[pagebackref=true,breaklinks=true,letterpaper=true,colorlinks,bookmarks=false]{hyperref}
\DeclarePairedDelimiterX{\infdivx}[2]{(}{)}{%
	#1\|#2%
}
\newcommand{\etal}{\textit{et al}.}
\newcommand{\ie}{\textit{i}.\textit{e}., }
\newcommand{\eg}{\textit{e}.\textit{g}. }

\begin{document}

%%%%%%%%% TITLE
\title{A Generative Map for Image-based Camera Localization}

\author{
\begin{tabular}{c}
	Mingpan Guo\\
	fortiss GmbH\\
	Munich, Germany\\
	{\tt\small m.guo@tum.de}\\
	\\
	Jiaojiao Ye\\
	Technical University of Munich\\
	Munich, Germany\\
	{\tt\small jiaojiao.ye@tum.de}
\end{tabular}
\begin{tabular}{c}
	Stefan Matthes\\
	fortiss GmbH\\
	Munich, Germany\\
	{\tt\small matthes@fortiss.org}\\
	\\
	Hao Shen\\
	fortiss GmbH\\
	Munich, Germany\\
	{\tt\small shen@fortiss.org}
\end{tabular}
}

\maketitle
%\thispagestyle{empty}

%%%%%%%%% ABSTRACT
\begin{abstract}
	In image-based camera localization systems, information about the environment is usually stored in some representation, which can be referred to as a map.
	Conventionally, most maps are built upon hand-crafted features.
	Recently, neural networks have attracted attention as a data-driven map representation, and have shown promising results in visual localization.
	However, these neural network maps are generally hard to interpret by human.
	A readable map is not only accessible to humans, but also provides a way to be verified when the ground truth pose is unavailable.
	To tackle this problem, we propose Generative Map, a new framework for learning human-readable neural network maps, by combining a generative model with the Kalman filter, which also allows it to incorporate additional sensor information such as stereo visual odometry.
	For evaluation, we use real world images from the 7-Scenes and Oxford RobotCar datasets.
	We demonstrate that our Generative Map can be queried with a pose of interest from the test sequence to predict an image, which closely resembles the true scene. 
	For localization, we show that Generative Map achieves comparable performance with current regression models.
	Moreover, our framework is trained completely from scratch, unlike regression models which rely on large ImageNet pretrained networks.
	\footnote{The code of this work can be found here: \url{https://github.com/Mingpan/generative_map}}
\end{abstract}

%%%%%%%%% BODY TEXT
\section{Introduction}
\begin{figure}
	\begin{center}
		\includegraphics[width=.5\linewidth]{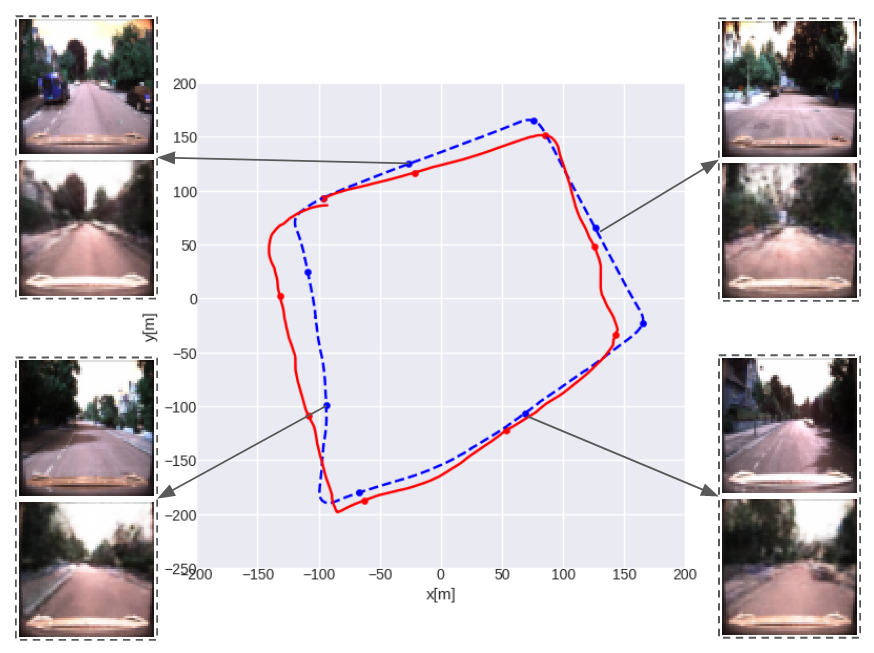}
	\end{center}
	\caption{Image generated from different poses taken from one trajectory of RobotCar with a driving length of $1120m$. The ground truth trajectory is showed in dashed blue, while the localization from generative map with stereo Visual Odometry (VO) is shown in solid red. We take four equidistant timestamps from the sequence, and show the real image from the true camera pose (top) together with the generated image provided by our DNN map using its output pose (bottom).}
	\label{fig:robotcar_gen_traj}
\end{figure}

Image-based localization is an important task for many computer vision applications, such as autonomous driving, indoor navigation, and augmented or virtual reality.
In these applications, the environment is usually represented by a map, whereby the approaches differ considerably in the way the map is structured.
In classical approaches, human designed features are extracted from images, and stored into a map with geometrical relations.
The same features can then be compared with the recorded ones to determine the camera pose relative to the map. 
Typical examples of these features include local point-like features~\cite{klein2007feature, murartal2015feature}, image patches~\cite{newcombe2011dtam, engel2014lsd}, and objects~\cite{salas2013feature}.

However, these approaches may ignore useful information that is not captured by the employed features.
This becomes more problematic if there are not enough rich textures to be extracted from the environment.
Furthermore, these approaches typically rely on prescribed structures like point clouds or grids, which are inflexible and grow with the scale of the environment.

Recently, deep neural networks (DNNs) are considered for the direct prediction of 6-DoF camera poses from images~\cite{kendall2015posenet, melekhov2017image, clark2017vidloc, walch2017image, brahmbhatt2018geometry}.
In this context, Brahmbhatt  ~\cite{brahmbhatt2018geometry} proposed to treat a neural network as a form of map representation, \ie an abstract summary of input data, which can be queried to get camera poses.
The DNN is trained to establish a relationship between images and corresponding poses.
During test time, it can be used for querying a pose given an input image from that viewpoint.
While the performance of these DNN map approaches has significantly improved~\cite{kendall2015posenet, kendall2017geometric, brahmbhatt2018geometry} and is getting close to hybrid approaches, \eg ~\cite{brachmann2017dsac}, these maps are typically unreadable for humans.

To solve this problem, we propose a new framework for learning a DNN map, which not only can be used for localization, but also allows queries from the other direction, \ie given a camera pose, what should the scene look like?
We achieve this via a combination of Variational Auto-Encoders (VAEs)~\cite{kingma2013auto} with a new training objective that is appropriate for this task, and the classic Kalman filter~\cite{kalman1960new}.
This makes the map human readable, and hence easier to interpret and verify.

Most research on image generation~\cite{vedantam2017generative, suzuki2018improving, gulrajani2016pixelvae, higgins2018scan, tolstikhin2017wasserstein, gregor2015draw} are either based on VAEs~\cite{kingma2013auto}, or Generative Adversarial Networks~\cite{goodfellow2014generative}. 
In our work, we take the VAE approach, due to its capability to infer latent variables from input images.
On the other hand, our model relies on the Kalman filter for connecting the sequence with a neural network as the observation model.
This also enables our framework to integrate the transition model of the system, and other sources of sensor information, if they are available.

Our main contributions are summarized as follows:
\begin{itemize}
	\item 
	Most prior works on DNN maps~\cite{kendall2015posenet, melekhov2017image, clark2017vidloc, walch2017image, brahmbhatt2018geometry} learn the map representation by directly regressing the 6-DoF camera poses from images.
	In this work, we approach this problem from the opposite direction via a generative model, \ie by learning the mapping from poses to images.
	For maintaining the discriminability, we derive a new training objective, which allows the model to learn pose-specific distributions, instead of a single distribution for the entire dataset as traditional VAEs~\cite{kingma2013auto, suzuki2018improving, vedantam2017generative, wang2016deep, higgins2018scan, tolstikhin2017wasserstein}.
	Our map is thus more interpretable, as it can be used for predicting an image from a particular viewpoint.
	
	\item 
	Generative models cannot directly produce poses for localization. 
	To solve this, we exploit the sequential structure of the localization problem, and propose a framework to \textit{estimate} the poses with a Kalman filter~\cite{kalman1960new}, where a neural network is used for the observation model of the filtering process.
	We show that this estimation framework works even with a simple constant transition model, is robust against large initial deviations, and can be further improved if additional sensor information is available.
	While being trained completely from scratch, it achieves comparable localization performance to the current regression based approaches~\cite{kendall2017geometric, walch2017image}, which rely on pretraining on ImageNet~\cite{imagenet_cvpr09}.
	
\end{itemize}

%-------------------------------------------------------------------------
\section{Related Works}

% Our work is inspired by recent research in DNN based visual SLAM, and image generation.

% Image based localization work like classical SLAM, SfM etc.

\paragraph{DNN map for localization}
In terms of localization, PoseNet~\cite{kendall2015posenet} first proposed to directly learn the mapping from images to the 6-DoF camera poses.
Follow-up works in this direction improved the localization performance by introducing deeper architectures~\cite{melekhov2017image}, exploiting spatial~\cite{clark2017vidloc} and temporal~\cite{walch2017image} structures, and incorporating relative distances between poses in the training objective~\cite{brahmbhatt2018geometry}.
Kendall and Cipolla~\cite{kendall2017geometric} showed that the idea of probabilistic deep learning can be applied, and introduced learnable weights for the translation and rotation error in the loss function, which increased the performance significantly.
All of these approaches tackle the learning problem via direct regression of camera poses from images, and focus on improving the accuracy of localization.
Instead, we propose to learn the generative process from poses to images.
%
%Regarding the full SLAM scenario, Henriques and Vedaldi~\cite{henriques2018mapnet} proposed an end-to-end DNN architecture with localization and map building capabilities.
%%
%However, their model is not stable and suffers from drifting problem even with static, simulated scenes.
%
Our focus is to make the DNN map human readable, by providing the capability to query the view of a specific pose.

\paragraph{Image Generation}
Generative models based on neural networks were originally designed to capture the image distribution~\cite{vincent2008auto, kingma2013auto, rezende2014auto, goodfellow2014generative, gregor2015draw}.
Recent works in this direction succeeded in generating images of increasingly higher quality.
However, these models do not establish geometric relationships between viewpoints and images.
In terms of conditional generation of images, many approaches have been proposed for different sources of information, \eg class labels~\cite{radford2015unsupervised} and attributes~\cite{vedantam2017generative, suzuki2018improving}. 
For a map in camera localization, our input source is the camera pose.
The generative query network~\cite{eslami2018neural} can generate images from different poses, for a variation of environments.
The follow-up work~\cite{rosenbaum2018learning} in this direction also aims at solving the localization problem with generative models.
However, their approach was only evaluated in simulated environments, and did not provide comparison with current regression based models.
Instead, we train and evaluate our framework on localization benchmark datasets with real images~\cite{shotton2013scene, RobotCarDatasetIJRR}.
% (TODO) s, both for indoor and outdoor environments.

\paragraph{VAE-based training objective}
Several recent works~\cite{suzuki2018improving, vedantam2017generative, wang2016deep, higgins2018scan, tolstikhin2017wasserstein} discuss VAE-based image generation.
Most of them assume a single normal distribution as the prior for the latent representation, and regularize the latent variable of each data point to match this prior~\cite{suzuki2018improving, vedantam2017generative, wang2016deep, higgins2018scan, yoo2017variational}.
Tolstikhin \etal~\cite{tolstikhin2017wasserstein} relaxed this constraint by modeling the latent representations of the entire dataset, instead of a single data point, as one single distribution.
However, such a setting is still inappropriate in our case, since restricting latent representations from different pose-image pairs to form a single distribution may reduce their discriminability, which can be critical for localization tasks.
There have been also several works proposed for sequence learning with VAEs~\cite{krishnan2015deep, yoo2017variational, karl2016deep} and sequential control problems~\cite{watter2015embed}, which similarly assume a single prior distribution for the latent variables.

Conditional VAE~\cite{sohn2015learning} can avoid this problem by conditioning the encoding and decoding processes with attributes.
However, when conditioning the decoder on attributes, the latent representation does not need to contain any information about the attributes, and cannot be used for inferring the attributes (poses) in our case.
Instead, we derive our training objective with pose specific latent representations, while avoiding conditioning the decoder on the poses.
By assuming each pose specific latent distribution to be Gaussian, we also make our proposed approach naturally compatible with a Kalman filter.
This is explained further in Section~\ref{sec:objective} and~\ref{sec:KF}.

%-------------------------------------------------------------------------
\section{Proposed Approach}

\begin{figure}
	\begin{center}
		\includegraphics[width=.5\linewidth]{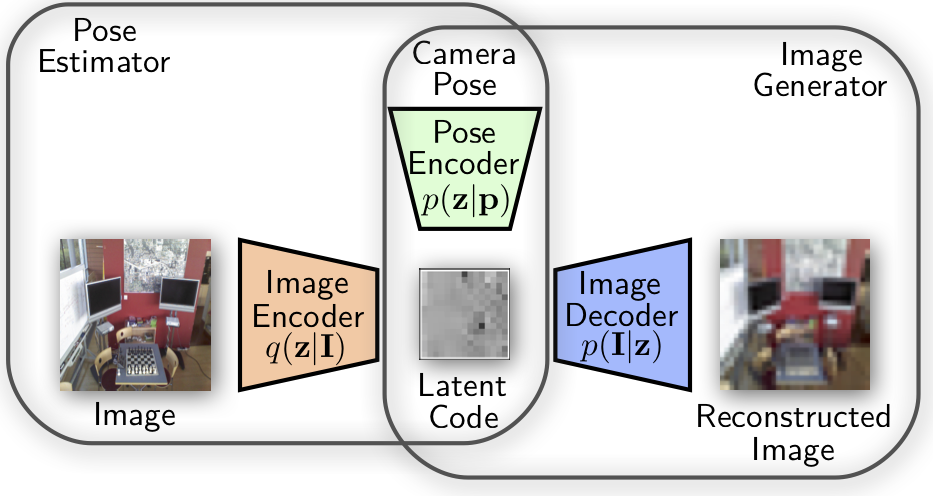}
	\end{center}
	\caption{Proposed architecture of Generative Map, where relevant networks for estimation and generation are shown as trapezoids.}
	\label{fig:model}
\end{figure}

In this paper, we propose a new framework for learning a DNN-based map representation, by learning a generative model.
Figure~\ref{fig:model} shows our overall framework, which is described in detail in Section~\ref{sec:framework}.
Our objective function is based on the lower-bound of the conditional log-likelihood of images given poses. 
In Section~\ref{sec:objective} we derive this objective for training the entire framework from scratch.
Section~\ref{sec:KF} introduces the pose estimation process for our framework.
The sequential estimator based on the Kalman filter is crucial for the localization task in our model, and allows us to incorporate the transition model of the system in a principled way.

In this work, we denote images by $\pmb{I}$, poses by $\pmb{p}$, and latent variables by $\pmb{z}$.
We assume the generative process $\pmb{p} \rightarrow \pmb{z} \rightarrow \pmb{I}$, and follow~\cite{kingma2013auto} to use $p(\cdot)$ and $q(\cdot)$ for generative and inference models, accordingly. 

\subsection{Framework}
\label{sec:framework}

Our framework consists of three neural networks, the image encoder $q (\pmb{z}|\pmb{I})$, pose encoder $p (\pmb{z}|\pmb{p})$, and image decoder $p(\pmb{I}|\pmb{z})$, as shown in Figure~\ref{fig:model}.
During training, all three networks are trained jointly with a single objective function described in Section~\ref{sec:objective}.
Once trained, depending on the task that we want to perform, \ie pose estimation or image (video) generation, different networks should be used.
Generating images involves the pose encoder $p(\pmb{z} | \pmb{p})$ and image decoder $p(\pmb{I} | \pmb{z})$, while pose estimation requires the pose encoder $p(\pmb{z} | \pmb{p})$ and image encoder $q(\pmb{z} | \pmb{I})$.

\subsection{Training Objective}
\label{sec:objective}
Our objective function is based on the Variational Auto-Encoder, which optimizes the following lower bound of the log-likelihood~\cite{kingma2013auto}
\begin{align}
\label{eq:vae}
\!\!\log p(x)
\!\geq\!& - D_{KL} \infdivx{q(z | x)}{p(z)} \! + \! \mathbb{E}_{q(z | x)} \! \left[\log p(x | z)\right],\!
\end{align}
where $x$ represents the data to encode, and $z$ stands for the latent variables that can be inferred by $x$ through $q(z|x)$. 
The objective can be intuitively interpreted as minimizing the reconstruction error $\mathbb{E}_{q(z | x)} \left[- \log p(x | z)\right]$ together with a KL-divergence term for regularization $D_{KL} \infdivx{q(z | x)}{p(z)}$.

To apply VAEs in cases with more than one input data source, \eg images $\pmb{I}$ and poses $\pmb{p}$ like in our case, we need to reformulate the above lower bound. 
We achieve this by optimizing the following lower bound,
\begin{align}
\label{eq:generator_bound}
\log p(\pmb{I}|\pmb{p}) 
=& \int q (\pmb{z}|\pmb{I}) \log p(\pmb{I}|\pmb{p}) d\pmb{z}\\\nonumber
=& \int q (\pmb{z}|\pmb{I}) \log p(\pmb{I}, \pmb{z}|\pmb{p}) d\pmb{z} 
- \int q (\pmb{z}|\pmb{I}) \log p(\pmb{z}|\pmb{p}, \pmb{I}) d\pmb{z}\\\nonumber
=& D_{KL} \infdivx{q (\pmb{z}|\pmb{I})}{p (\pmb{z}|\pmb{p}, \pmb{I})} 
- D_{KL} \infdivx{q (\pmb{z}|\pmb{I})}{p (\pmb{z}|\pmb{p})} + \mathbb{E}_{q (\pmb{z}|\pmb{I})} \left[\log p(\pmb{I}|\pmb{z})\right] \\\nonumber
\geq& -D_{KL} \infdivx{q (\pmb{z}|\pmb{I})}{p (\pmb{z}|\pmb{p})} + \mathbb{E}_{q (\pmb{z}|\pmb{I})} \left[\log p(\pmb{I}|\pmb{z})\right].
\end{align}
%
% where $ q (\pmb{z}|\pmb{I}) $, $ p (\pmb{z}|\pmb{p}) $ and $ p(\pmb{I}|\pmb{z}) $ are modeled by DNNs. 
%
% A detailed derivation can be found in the appendix.
%
For convenience, we treat the negative lower bound as our loss and train our model by minimizing
\begin{align}
\label{eq:generator}
\mathcal{L}
=& \mathbb{E}_{q (\pmb{z}|\pmb{I})} \left[- \log p(\pmb{I}|\pmb{z})\right] + D_{KL} \infdivx{q (\pmb{z}|\pmb{I})}{p (\pmb{z}|\pmb{p})}.
\end{align}

Similar to Equation~\eqref{eq:vae}, the first term in our loss function can be seen as a reconstruction error for the image, while the second term serves as a regularizer. 
Unlike most other extensions of the VAE \cite{kingma2013auto, vedantam2017generative, wang2016deep}, where the marginal distribution of latent variable $\pmb{z}$ is assumed to be normally distributed, our loss function assumes the distribution of latent variables to be normal, only when conditioning on the corresponding poses or images.
We assume that for every pose $\pmb{p}$, the latent representation follows a normal distribution $\mathcal{N}(\pmb{\mu_{z | p}}, \pmb{\Sigma_{z | p}})$. 
Similarly, a normal distribution is assumed for the latent variable conditioning on the image $\mathcal{N}(\pmb{\mu_{z | I}}, \pmb{\Sigma_{z | I}})$.
The KL-divergence term enforces these two distributions to be close to each other.

One fundamental difference between our loss function~\eqref{eq:generator} and previous works in DNN-based visual localization~\cite{kendall2015posenet, kendall2017geometric, brahmbhatt2018geometry} is, a direct mapping from images to poses does not exist in our framework.
Hence, we cannot obtain the poses by direct regression.
Instead, we treat the network $p(\pmb{z} | \pmb{p})$ as an observation model and use the Kalman filter~\cite{kalman1960new} for iteratively estimating the correct pose. 
This is described in detail in Section~\ref{sec:KF}.
Another important difference is that the generative process from poses to images is modeled by the networks $p(\pmb{z} | \pmb{p})$ and $ p(\pmb{I}|\pmb{z}) $.
This allows us to query the model with a pose of interest, and obtain a generated RGB image which describes how the scene should look like at that pose.

\subsection{Kalman Filter for Pose Estimation}
\label{sec:KF}

\begin{figure}
	\begin{center}
		\includegraphics[width=.6\linewidth]{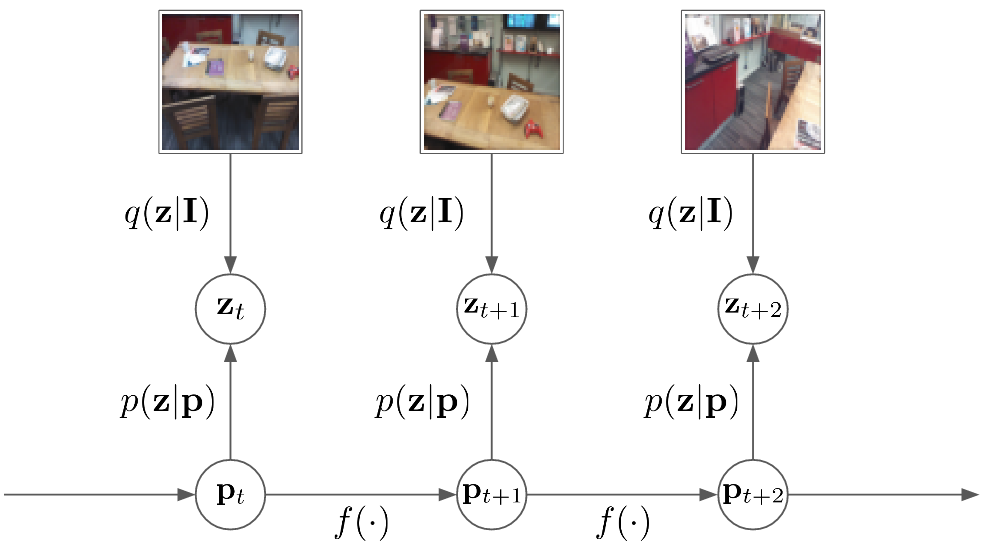}
	\end{center}
	\caption{Update of the Kalman filter for pose estimation. Pose $\pmb{p}$ and latent variable $\pmb{z}$ are seen as the state and observation, respectively.}
	\label{fig:KF}
\end{figure}

As mentioned above, the generative model we propose cannot predict poses directly. 
However, we can still estimate the pose with the trained model using a Kalman filter, as shown in Figure~\ref{fig:KF}. 
The network $q(\pmb{z} | \pmb{I})$ is seen as a sensor, which processes an image at each time step, and produces an observation vector $\pmb{z}$ based on that image $\pmb{I}$.
From the pose we can also obtain an expected observation using $p(\pmb{z} | \pmb{p})$, which is compared with the observation from the raw image.
By assumption, $q(\pmb{z} | \pmb{I})$ and $p(\pmb{z} | \pmb{p})$ are both normally distributed and regularized to resemble each other.
In addition, we also model $\pmb{p}$ as normally distributed.
Therefore, the generator model naturally fits into the estimation process of a Kalman filter.

In order to close the update loop, we need a transition function $\pmb{p}_{t+1} = f(\pmb{p}_t)$. 
If the ego-motion is unknown, a simple approach is to assume the pose remains constant over time, \ie $\pmb{p}_{t+1} = \pmb{p}_t$.
In such a case, the Kalman filter introduces no further information about the system itself, but rather a smoothing effect based on previous inputs.
If additional control signals or motion constraints are known, a more sophisticated transition model can be devised.
In such a case, the transition function becomes $\pmb{p}_{t+1} = f(\pmb{p}_{t}, \pmb{u}_{t})$, where $\pmb{u}_t$ is the control signal for the ego-motion, which can be obtained from other sensors.
%%
%We show in Section~\ref{exp:7scenes_loc} that we can estimate the pose with a simple constant transition model.
%%
%And a significant improvement in localization performance can be achieved, if an accurate transition model from $\pmb{p}_{t}$ to $\pmb{p}_{t+1}$ is available.

\begin{table}
	\begin{center}
		\begin{tabular}{|c|c|}
			\hline
			Training Framework & Kalman Filter  \\
			\hline\hline
			Pose $\pmb{p}$ & State \\
			\hline
			Mean of & \\
			latent variable $\mu_z$ & Observation\\
			\hline
			Variance of &  Diagonal of observation\\
			latent variable $\sigma_z$ & uncertainty matrix $\pmb{R}$ \\
			\hline
			& Sensor that \\
			Image encoder $q(\pmb{z} | \pmb{I})$ & produces observation and $\pmb{R}$\\
			\hline
			Pose encoder $p(\pmb{z} | \pmb{p})$ & Observation model $g(\pmb{p})$ \\
			\hline
		\end{tabular}
	\end{center}
	\caption{Corresponding relationship between different components of our training framework, and the Kalman filter during pose estimation. Note that, instead of sampling from $q(\pmb{z} | \pmb{I})$, we directly use the mean $\mu_z$ as the observation for the Kalman filter, which increases the stability. }
	\label{tab:KF}
\end{table}

The corresponding relationship between different components in our training framework and Kalman filter is summarized in Table~\ref{tab:KF}.
The pose estimation update using the Kalman filter consists of \textit{prediction} and \textit{correction} step, which are explained in the following.
\begin{itemize}
	\item \textbf{Prediction with transition model}\\
	Let us denote the transition function by $f(\pmb{p}, \pmb{u})$, and its first order derivative w.r.t. $\pmb{p}$ by $f_p$, which can be estimated by the finite difference method.
	An update for the prediction step of the estimation can then be written as
	\begin{align}
	\label{eq:KF_predict1}
	\pmb{p} \leftarrow& f(\pmb{p}, \pmb{u}) \\
	\label{eq:KF_predict2}
	\pmb{\Sigma}_p \leftarrow& f_p \pmb{\Sigma}_p f_p^T + \pmb{Q},
	\end{align}
	where $\pmb{\Sigma}_p$ stands for the covariance matrix of the pose, and $\pmb{Q}$ for the state transition uncertainty. 
	It needs to be set to higher values if the transition is inaccurate, \eg if we are using a constant model, and smaller when an accurate transition model is available.
	
	\item \textbf{Correction with current observation}\\
	We denote the neural observation model $p(\pmb{z} | \pmb{p})$ by $g(\pmb{p})$, its first order derivative w.r.t. $\pmb{p}$ given by the finite difference method is denoted by $g_p$.
	In each time step, our neural sensor model $q(\pmb{z} | \pmb{I})$ produces a new observation $\mu_z$ based on the current image $\pmb{I}$.
	The correction step can then be written as
	\begin{align}
	\pmb{K} \leftarrow& \pmb{\Sigma}_p g_p^T 
	\left(g_p \pmb{\Sigma}_p g_p^T + \pmb{R}\right)^{-1} \\
	\pmb{p} \leftarrow& \pmb{p} + \pmb{K} (\mu_z - g(\pmb{p})) \\
	\pmb{\Sigma}_p \leftarrow& \pmb{\Sigma}_p - \pmb{K} g_p \pmb{\Sigma}_p,
	\end{align}
	where $\pmb{K}$ is the Kalman gain, and $\pmb{R}$ is the observation uncertainty.
	In our case, we can directly use the variance of $\pmb{z}$ inferred by the image encoder $q(\pmb{z} | \pmb{I})$ to build the matrix $\pmb{R}$.
	
\end{itemize}

\subsection{Implementation}
\label{sec:implementation}

We use DC-GAN~\cite{radford2015unsupervised} with 512 initial feature channels for both image encoder $q(\pmb{z}|\pmb{I})$ and decoder $p(\pmb{I}|\pmb{z})$. 
The dimension of the latent variable $z$ is set as $128$ for 7-Scenes and $256$ for RobotCar.
For the pose encoder, we use a standard 3-layer fully connected network, where the only hidden layer contains 512 units.
Our generative map is trained from scratch without any pretrained model from ImageNet, unlike regression based approaches \eg PoseNet~\cite{kendall2015posenet, kendall2017geometric} and Pose-LSTM~\cite{walch2017image}.
%%
%Training our model on both 7-Scenes and robotcar datasets (less than 10,000 images per scene)
%%
%\footnote{This training set size is also significantly smaller than usual generative learning datasets. For example, CelebA~\cite{liu2015faceattributes} is a popular dataset used by generative models, which has more than 200,000 images.}
%%
%doesn't result in obvious overfitting.
%%
More details about the architecture setup will be provided in the supplementary material.

The input images for the image encoder are resized to $96\times 96$, while generated images are set to be $64\times 64$. 
The input poses for the pose encoder are normalized to achieve same scale for translation and rotational coordinates.
During training, the variance $\sigma_{z|p}$ is fixed to $1$ to better restrict the latent representation span.
For modeling the Gaussian distribution in image reconstruction, we tested different values for the variance and find $e^{-3}$ performs the best.
We use the Adam optimizer~\cite{kingma2014adam} with a learning rate of 0.0001 without decay.
The model is trained for 5,000 epochs on each environment from 7-Scenes, and 3,000 epochs on RobotCar.

% For comparison, we implemented PoseNet-2017~\cite{kendall2017geometric} with ResNet-34 pretrained from ImageNet, according to the description given by~\cite{brahmbhatt2018geometry}, as it is to our knowledge the best performing PoseNet architecture.
%
% Unlike in our model, we use a much higher input resolution of $197\times 197$ for PoseNet-2017, as ResNet-34 requires it as the minimum acceptable input size.
%
% Following~\cite{brahmbhatt2018geometry}, we train 300 epochs
%
% \footnote{The difference in the training epochs might seem significant, but pretraining on ImageNet takes much more effort in terms of computation, which is necessary for regression models like PoseNet.}
%
% for PoseNet.

%-------------------------------------------------------------------------
\section{Experiments}

\begin{figure}
	\begin{center}
		\includegraphics[width=.8\linewidth]{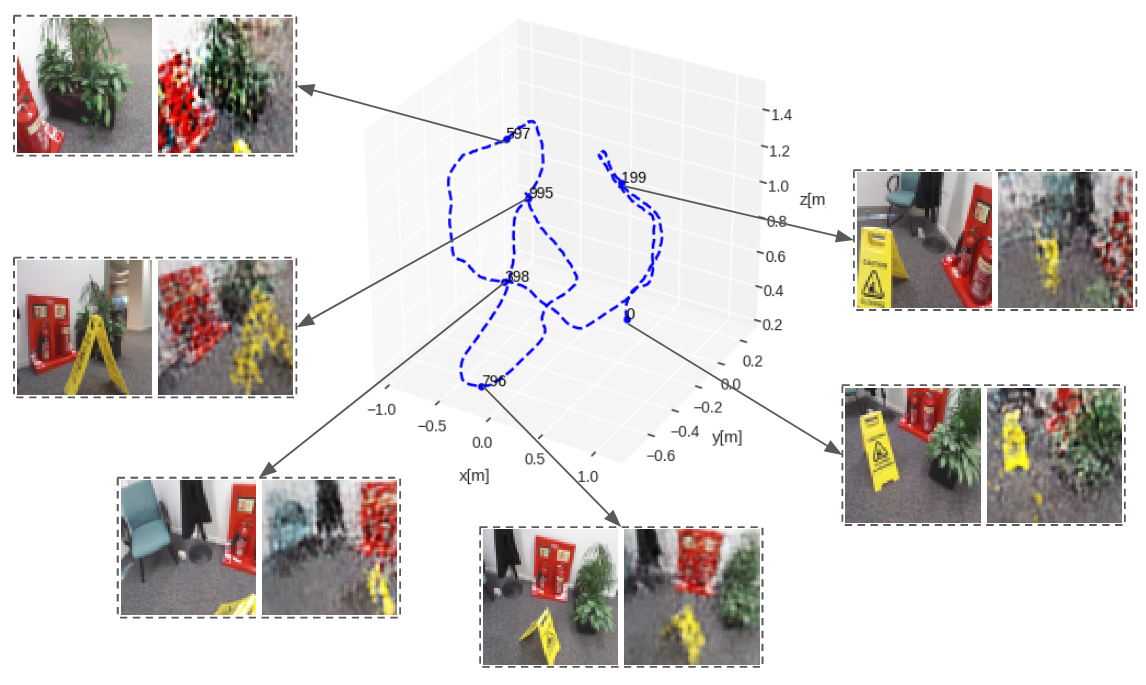}
	\end{center}
	\caption{Image generated from different poses taken from the scene ``fire''. The camera trajectory and its corresponding poses are not observed during training. We take six equidistant timestamps (numbers in black) from the sequence, and show the real image from that camera pose (left) together with the generated image provided by our DNN map (right).}
	\label{fig:7scene_gen_traj}
\end{figure}

We use the 7-Scenes~\cite{shotton2013scene} and Oxford RobotCar~\cite{RobotCarDatasetIJRR} datasets to evaluate our framework, for both generation and localization tasks.
The 7-Scenes~\cite{shotton2013scene} dataset contains video sequences recorded from seven different indoor environments.
Each scene contains two to seven sequences for training, with either 500 or 1000 images for each sequence.
The corresponding ground truth poses are provided for training and evaluation.
Oxford RobotCar~\cite{RobotCarDatasetIJRR} provides a large dataset with not only images, but also LIDAR, GPS, INS, and stereo Visual Odometry (VO) collected from camera and sensors mounted on a car.
We extract one subset from the RobotCar with a total driving length of $1120m$, which was also used in~\cite{brahmbhatt2018geometry}.
We follow their setup for dividing training and test sequences.
% For training generative models, prior approaches often rely on a large dataset, \eg CelebA~\cite{liu2015faceattributes} contains more than 200,000 images. 
%
% Therefore, the dataset we use is much more challenging, where each scene only contains less than 10,000 training samples.

% TODO: what we do in each section?

\subsection{Generation for Map Reading}

\begin{figure*}
	\begin{center}
		\includegraphics[width=.8\linewidth]{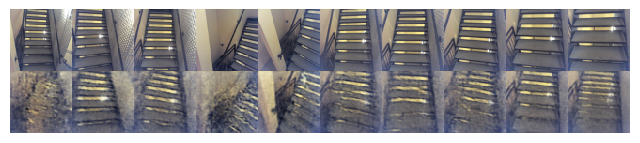}
		\includegraphics[width=.8\linewidth]{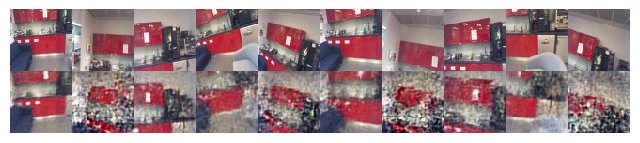}
		\includegraphics[width=.8\linewidth]{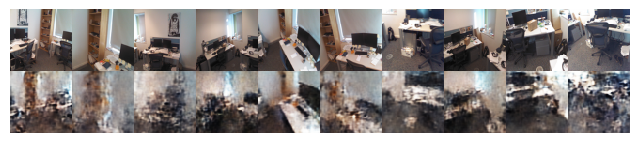}
		\includegraphics[width=.8\linewidth]{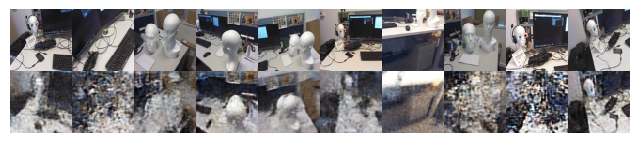}
		\includegraphics[width=.8\linewidth]{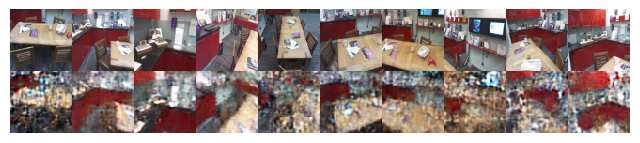}
		\includegraphics[width=.8\linewidth]{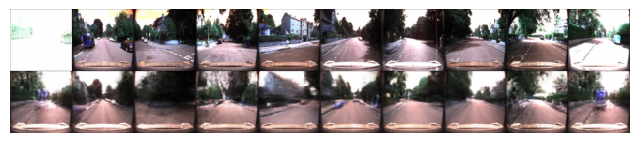}
	\end{center}
	\caption{Image generation for equidistant poses from the test sequences. Scenes from top to bottom: stairs, pumpkin, office, 
		%kitchen, 
		heads, kitchen,
		and RobotCar. For each sequence, the down-sampled original images are shown on top, and the generated images on bottom. }
	\label{fig:7scene_generation}
\end{figure*}

The generation capability enables us to read the DNN map by querying for a specific pose, even when the network might not have encountered that pose during training. 
In particular, we provide the pose of interest to the pose encoder $p(\pmb{z}|\pmb{p})$ and obtain a latent representation $\pmb{z}$ that describes the scene in a high dimensional space. 
This latent representation is then passed to the image decoder $p(\pmb{I}|\pmb{z})$ for generating a human-readable image, which shows how the scene should look like from that particular pose of interest.

Figure~\ref{fig:7scene_gen_traj} shows generated images of poses taken from equidistant sample of one unseen test sequence, from the indoor environment ``fire'' in the 7-Scenes dataset. 
From the result we can observe that the generative map is able to predict plausible images for different queried poses.
Main objects of each scene can be observed in each queried image, in their corresponding positions.
Generated images for other indoor scenes and the RobotCar are also shown in Figure~\ref{fig:7scene_generation}, which exhibit similar results.
Interestingly, there are some timesteps in the RobotCar datasets when the original picture is over exposed, but our generative map is robust against these individual corrupted training samples, and still able to predict reasonable scenes for these poses.
This experiment shows that the generative map has successfully captured the essential information of the environments, which is necessary for it to be used for localization.

\subsection{Localization}

To evaluate the localization performance of our framework, we provide the model with a sequence of camera images, and utilize the Kalman filter as described in Section~\ref{sec:KF} to iteratively estimate the current pose of the camera.
Such iterative estimation requires a predefined state uncertainty matrix $\pmb{Q}$ as in Equation~\eqref{eq:KF_predict2}, a given state transition model $f(\cdot)$ as in Equation~\eqref{eq:KF_predict1}, and an initial condition $\pmb{p}_0, \pmb{\Sigma}_{p_0}$ as the starting point.
We conduct extensive experiments that study the impacts of these three components in this section.
Specifically, Section~\ref{sec:localize_constant} evaluates the model performance with a constant transition model, and Section~\ref{sec:localize_accurate} provides experimental results with a more accurate model.

For convenience, we only consider diagonal matrices for $\pmb{\Sigma}_{p_0}$ and $\pmb{Q}$ with identical values in their diagonals.
Furthermore, we assume $\pmb{Q} = \pmb{\Sigma}_{p_0}$. 
In this case, these two matrices can be identified with a single scalar value $\sigma_q$, \ie $\pmb{\Sigma}_{p_0} = \pmb{Q} = \sigma_q I$.

\subsubsection{Constant Transition Model}
\label{sec:localize_constant}

The transition function $f(\cdot)$ directly influences the accuracy of the prediction step in our Kalman Filter (Equation~\eqref{eq:KF_predict1}).
However, an accurate transition model is not always available, and sometimes we need to settle for a constant model as an alternative, as described in Section~\ref{sec:KF}.
Here we first evaluate the localization performance using a constant model for different $\sigma_q$, as shown in Table~\ref{tab:constant}.
From the localization results we can observe that, when $\sigma_q \geq 10^{-3}$, the smoothing effect introduced by the Kalman filter is almost negligible.
However, from $\sigma_q = 10^{-3}$ to $\sigma_q = 10^{-6}$, the smoothing effect becomes too strong, and decreases the localization performance.
Based on the above result, we select $\sigma_q = 10^{-3}$ as our default setting for 7-Scenes.
\begin{table}
	\small
	\begin{center}
		\begin{tabular}{c|c|c|c}
			Scene & $\sigma_q = 10^0$ & $\sigma_q = 10^{-3}$ & $\sigma_q = 10^{-6}$ \\
			\hline
			Chess &$0.18m$, $6.35^\circ$&$0.18m$, $6.32^\circ$&$0.36m$, $12.33^\circ$ \\
			Fire &$0.31m$, $12.50^\circ$&$0.31m$, $12.47^\circ$&$0.39m$, $17.04^\circ$ \\
			Heads &$0.26m$, $22.63^\circ$&$0.24m$, $20.15^\circ$&$0.38m$, $28.31^\circ$ \\
			Office &$0.38m$, $10.00^\circ$&$0.37m$, $9.77^\circ$&$0.46m$, $13.80^\circ$ \\
			Pumpkin &$0.35m$, $7.13^\circ$&$0.34m$, $7.16^\circ$&$0.53m$, $10.83^\circ$ \\
			Kitchen &$0.50m$, $10.50^\circ$&$0.49m$, $10.42^\circ$&$0.60m$, $14.29^\circ$ \\
			Stairs &$0.50m$, $11.18^\circ$&$0.50m$, $11.28^\circ$&$0.58m$, $12.09^\circ$ \\
			\hline
			Average &$0.35m$, $11.47^\circ$&$0.35m$, $11.07^\circ$&$0.47m$, $15.53^\circ$ \\
		\end{tabular}
	\end{center}
	\caption{Localization error of generative map measured in meter ($m$) and degree ($
		^\circ$) with different state uncertainty $\sigma_q$, for the 7-Scenes dataset. We follow PoseNet-2017~\cite{kendall2017geometric} and report the median error. A constant transition model is applied, and the sequences are initialized with the correct starting poses. }
	\label{tab:constant}
\end{table}

\begin{figure}
	\begin{center}
		\includegraphics[width=.5\linewidth]{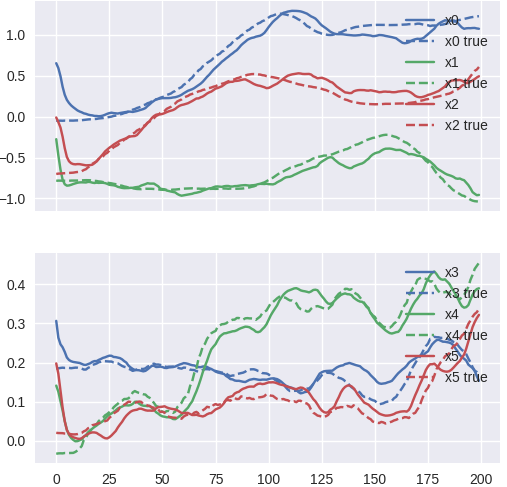}
	\end{center}
	\caption{Generative map for estimating the first test sequence from the 7-Scenes-chess environment. It is based on a constant transition model, with $\sigma_q = 10^{-4}$ and an initial deviation of $+1.0$. $x_0$ to $x_5$ in the picture represent the state, which is the concatenation of translation coordinates and log-quaternions~\cite{brahmbhatt2018geometry}. }
	\label{fig:deviation}
\end{figure}

The initial pose $\pmb{p}_0$ is also critical for a correct estimation, especially in the early timesteps.
Similar to the transition model, accurate initializations are sometimes also inaccessible.
Hence, it is interesting to study how the localization performs, not just with an accurate initialization, but also with an initial deviation.
In case of a constant transition model, the system needs to rely fully on the obtained images to correct the initial deviation and localize itself.
Table~\ref{tab:constant_deviation} shows the experiments in 7-Scenes with a constant model under different levels of initial deviation, for different state uncertainty $\sigma_q$.
The experiments show that even with a deviation as large as $\pm 1.0$, \ie $1.0m$ deviation in each direction, our map can still perform well.
Except when the state uncertainty $\sigma_q$ is too small, then the deviation causes larger error, as the model now needs more timesteps to correct itself.
One example of estimation with initial deviation is shown in Figure~\ref{fig:deviation}.

\begin{table*}
	\small
	\begin{center}
		\begin{tabular}{c | c|c | c|c | c|c }
			& 
			\multicolumn{2}{|c|}{\centering $\sigma_q = 10^{-3}$} &
			\multicolumn{2}{|c|}{\centering $\sigma_q = 10^{-4}$} & 
			\multicolumn{2}{|c}{\centering $\sigma_q = 10^{-5}$} \\
			abs. deviation & 
			$0.5$ & $1.0$ & 
			$0.5$ & $1.0$ &
			$0.5$ & $1.0$ \\
			\hline
			Chess &
			$0.18m, 6.33^\circ$&$0.18m, 6.34^\circ$&
			$0.19m, 6.41^\circ$&$0.18m, 6.43^\circ$&
			$0.23m, 7.69^\circ$&$0.23m, 7.77^\circ$\\
			Fire &
			$0.31m, 12.48^\circ$&$0.31m, 12.52^\circ$&
			$0.31m, 12.46^\circ$&$0.31m, 12.57^\circ$&
			$0.29m, 13.22^\circ$&$0.31m, 13.65^\circ$\\
			Heads &
			$0.24m, 20.15^\circ$&$0.24m, 20.32^\circ$&
			$0.26m, 18.06^\circ$&$0.26m, 19.91^\circ$&
			$0.27m, 31.58^\circ$&$0.28m, 30.70^\circ$\\
			Office &
			$0.37m, 9.76^\circ$&$0.37m, 9.79^\circ$&
			$0.37m, 9.77^\circ$&$0.37m, 9.81^\circ$&
			$0.35m, 9.08^\circ$&$0.36m, 9.13^\circ$\\
			Pumpkin &
			$0.34m, 7.25^\circ$&$0.35m, 7.62^\circ$&
			$0.40m, 8.18^\circ$&$0.45m, 8.51^\circ$&
			$0.60m, 12.57^\circ$&$0.64m, 13.60^\circ$\\
			Kitchen &
			$0.50m, 10.50^\circ$&$0.51m, 10.53^\circ$&
			$0.49m, 10.33^\circ$&$0.50m, 10.43^\circ$&
			$0.54m, 10.81^\circ$&$0.55m, 10.99^\circ$\\
			Stairs &
			$0.50m, 11.29^\circ$&$0.50m, 11.34^\circ$&
			$0.51m, 11.72^\circ$&$0.52m, 11.81^\circ$&
			$0.54m, 12.52^\circ$&$0.54m, 12.44^\circ$\\
			\hline
			Average &
			$0.35m, 11.11^\circ$&$0.35m, 11.21^\circ$&
			$0.36m, 10.99^\circ$&$0.37m, 11.35^\circ$&
			$0.40m, 13.91^\circ$&$0.42m, 14.04^\circ$\\		
		\end{tabular}
	\end{center}
	\caption{Initial deviation experiment for in the 7-Scenes dataset in meter ($m$) and degree ($
		^\circ$). For each $\sigma_q$, we introduce an initial deviation of $\pm 0.5$ and $\pm 1.0$ in each dimension, and average over each absolute deviation level to calculate the final error. For example, if the true initial pose is $x = y = z = 0$ with quaternion $q = [1, 0, 0, 0]$, the initial state will be a concatenation of translation coordinates with log-quaternion~\cite{brahmbhatt2018geometry}, \ie $[0, 0, 0, 0, 0, 0]$. After the deviation of $+ 1.0$, the state will become $[1, 1, 1, 1, 1, 1]$.}
	\label{tab:constant_deviation}
\end{table*}

We also find that the generative map performs comparably in localization with current regression based approaches, namely PoseNet2017~\cite{kendall2017geometric} and PoseLSTM~\cite{walch2017image}, as shown in Table~\ref{tab:constant_compare}.
Although the performance of our generative map does not consistently surpass that of the regression approaches, it can be trained completely from scratch without any ImageNet pretrained models. 
\begin{table}
	\small
	\begin{center}
		\begin{tabular}{c|c|c|c}
			&Generative Map & PoseLSTM & PoseNet17 \\
			&&\cite{walch2017image}&\cite{kendall2017geometric}\\
			\hline
			Chess &$0.18m$, $6.32^\circ$ &$0.24m$, $5.77^\circ$&$0.13m$, $4.48^\circ$ \\
			Fire &$0.31m$, $12.47^\circ$ &$0.34m$, $11.9^\circ$&$0.27m$, $11.30^\circ$\\
			Heads &$0.24m$, $20.15^\circ$&$0.21m$, $13.7^\circ$&$0.17m$, $13.00^\circ$ \\
			Office &$0.37m$, $9.77^\circ$&$0.30m$, $8.08^\circ$&$0.19m$, $5.55^\circ$ \\
			Pumpkin &$0.34m$, $7.16^\circ$&$0.33m$, $7.00^\circ$&$0.26m$, $4.75^\circ$ \\
			Kitchen &$0.49m$, $10.42^\circ$&$0.37m$, $8.83^\circ$&$0.23m$, $5.35^\circ$ \\
			Stairs &$0.50m$, $11.28^\circ$&$0.40m$, $13.7^\circ$&$0.35m$, $12.40^\circ$ \\
			\hline
			Average &$0.35m$, $11.07^\circ$&$0.31m$, $9.85^\circ$&$0.23m$, $8.12^\circ$ \\
		\end{tabular}
	\end{center}
	\caption{Comparison of median localization error between constant transition generative map and regression based approaches, measured in meter ($m$) and degree ($^\circ$). The generative map is trained from scratch, while the regression models are trained based on ImageNet pretrained networks. 
		% For each model, the input resolution is also stated.
		% For more comparison, we also implemented a lower input resolution version of PoseNet2017 on our own with ResNet34 pretrained on ImageNet. We reduce the input resolution of PoseNet2017 in our implementation as much as possible, as $197\times 197$ is the minimum required size for ResNet34. 
		% Note that the pretrained networks used by the regression models are different, and might result in difference in their performance. 
	}
	\label{tab:constant_compare}
\end{table}

\subsubsection{Accurate Transition Model}
\label{sec:localize_accurate}

\begin{figure*}
	\centering
	\begin{subfigure}[b]{0.19\textwidth}
		\includegraphics[width=1.\textwidth]{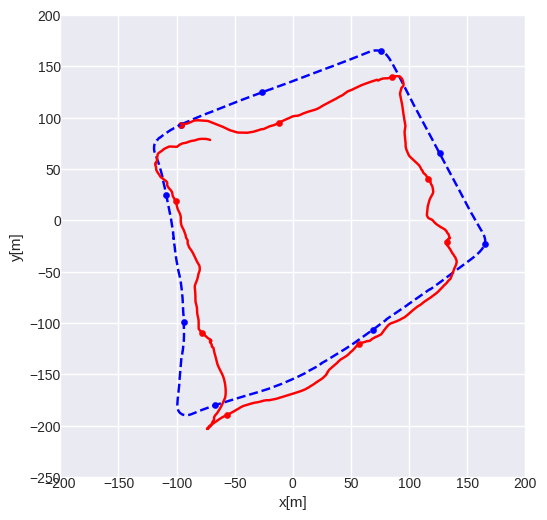}
		\caption{$\sigma_q = 10^{-17}$}
	\end{subfigure}
	\begin{subfigure}[b]{0.19\textwidth}
		\includegraphics[width=1.\textwidth]{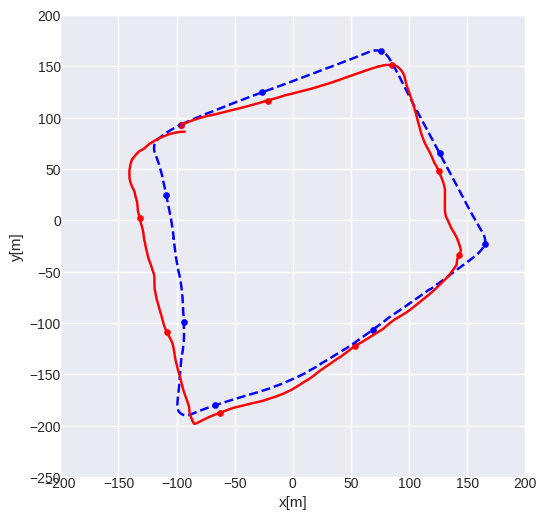}
		\caption{$\sigma_q = 10^{-18}$}
	\end{subfigure}
	\begin{subfigure}[b]{0.19\textwidth}
		\includegraphics[width=1.\textwidth]{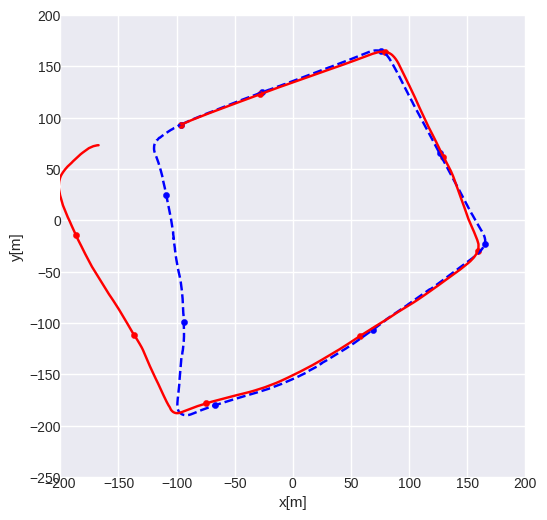}
		\caption{$\sigma_q = 10^{-19}$}
	\end{subfigure}
	\begin{subfigure}[b]{0.19\textwidth}
		\includegraphics[width=1.\textwidth]{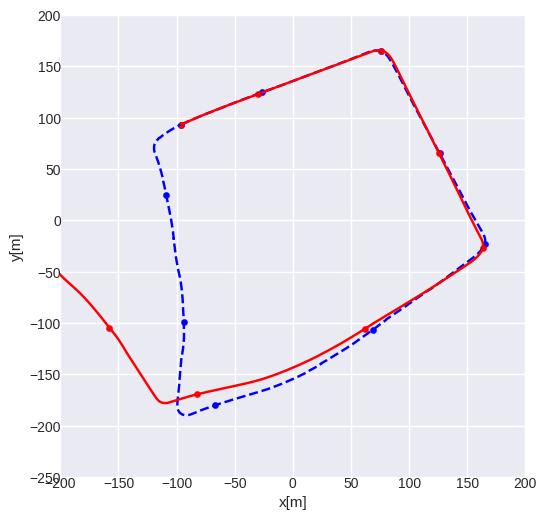}
		\caption{$\sigma_q = 10^{-30}$ (VO)}
	\end{subfigure}
	\caption{The localization performance on RobotCar for generative map with transition model given by the stereo VO. Results of different $\sigma_q$ are shown. Note that with $\sigma_q = 10^{-30}$ the measurements (images from the camera) are almost completely neglected, and the trajectory is just based on the stereo VO information.
	}
	\label{fig:robotcar_vo}
\end{figure*}

% When additional sensory data (\eg GPS, VO) are available, a more accurate transition model $f(\cdot)$ can be devised.
%
One advantage of applying Kalman filter is that it provides a principled way to incorporate transition models, if they are available.
Here we demonstrate the performance of our generative map with reasonable transition models.
In particular, we use difference-in-pose as the control signal to devise a simple transition model in \eqref{eq:KF_predict1}, \ie $\pmb{u} = \Delta p$.
Unlike in constant models, where the state uncertainty parameter $\sigma_q$ only serves as a way to connect and smooth sequential estimates, the $\sigma_q$ now defines how much the model relies on the measurements (images descriptions $\pmb{z}$ from $q(\pmb{z}|\pmb{I})$) and its transition control signal $\Delta p$.
In this subsection, we again explore the performance by providing different $\sigma_q$ to the framework.
%
% The ground truth initial poses are given for experiments in this section.

% In 7-Scenes, the images are the only sensory source, therefore we can only mimic an accurate transition model by feeding the true difference-in-pose $\Delta p^{true}$ to our framework.
%
% In other words, we are using the ground truth transition model.
%
In 7-Scenes, the images are the only information source, therefore we cannot devise a reasonable transition model without utilizing the ground truth values.
While in RobotCar, we can use the stereo VO to provide the control signal $\Delta p^{VO}$.
It is well-known that stereo VO information are more accurate locally, but might introduce drifts for a longer horizon.
This feature makes it appropriate to incorporate the stereo VO information as the per step transition model for our generative map.

The localization results of the generative map with stereo VO on RobotCar are shown in Figure~\ref{fig:robotcar_vo}.
An obvious trade-off is that a larger $\sigma_q$ allows the model to rely more on the image measurements, but less on the transition model.
If $\sigma_q$ is too large, the resulting trajectory might over-react to the input images and become instable.
In contrast, smaller $\sigma_q$ tells the model to rely less on the input images, but trust more on its inherent transition model.
At the extreme, $\sigma_q = 10^{-30}$ produces a trajectory that is almost identical to that of the original VO.
While the generative map with $\sigma_q = 10^{-18}$ effectively combines the information from the images and stereo VO, and produce a stabilized, self-corrected trajectory.

%%-------------------------------------------------------------------------
\section{Conclusion}
In image based localization problems, the map representation plays an important role.
Instead of using hand-crafted features, deep neural networks are recently explored as a way to learn a data-driven map~\cite{brahmbhatt2018geometry}.
Despite their success in improving the localization accuracy, prior works in this direction~\cite{kendall2015posenet, melekhov2017image, clark2017vidloc, walch2017image, kendall2017geometric, brahmbhatt2018geometry} produce maps that are unreadable for humans, and hence hard to visualize and verify.
In this work, we propose the \textit{Generative Map} framework for learning a human-readable neural network based map representation.
%%
%Our probabilistic framework tackles the problem of readability of prior DNN-based maps, by allowing queries for images given poses of interest.
%%
%Our training objective is derived from the generative model of Variational Auto-Encoders~\cite{kingma2013auto} and can be applied to train the entire framework jointly.
%%
%For localization, our approach relies on the classic Kalman filter~\cite{kalman1960new} and estimates the pose through an iterative process.
%%
Integrating Kalman filter into our model enables us to easily incorporate additional information, \eg sensor inputs of the system.

We evaluate our approach on the 7-Scenes~\cite{shotton2013scene} and RobotCar~\cite{RobotCarDatasetIJRR} dataset.
Our experimental result shows that given a pose of interest from the test data, our model is able to generate an image that largely matches the ground truth image from the same pose.
Moreover, we also show that our map can achieve comparable performance with the regression based approaches using only a constant transition model.
We also observe that, if other sensory data are available, \eg stereo VO in RobotCar, our generative map can effectively incorporate this information to the transition model, avoiding both instability from image measurements and drifting behavior from stereo VO.

The present work leads to several potential directions for future research.
First, the generated images may provide a way to visualize and measure the accuracy of the model for each region of the environment.
It is interesting to conduct an in-depth investigation regarding the correlation between the quality of the generated images and the localization accuracy.
Regions with worse generated images may require more training data to be collected.
%
% In that way, we are able to actively search for areas to improve based on \eg image reconstruction error from the environment.
%
Secondly, combining both generative and regression based DNN-map may result in a hybrid model with better readability and localization performance.
Finally, it is meaningful and interesting to extend our framework to a full SLAM scenario~\cite{henriques2018mapnet}, which can not only localize itself, but also build an explicit map in completely new environments.

{\small
\bibliographystyle{plain}
\bibliography{generative_map.bib}
}

\end{document}